\documentclass[runningheads]{llncs}

\usepackage{glossaries}
\usepackage[hidelinks,
            pdfauthor={Yochem van Rosmalen, Florian van der Steen, Sebastiaan Jans, Daan van der Weijden},
            pdftitle={Bursting the Burden Bubble},
            pdfsubject={Fairness metrics},
            pdfkeywords={Fairness metrics, Burden, Statistical parity, Decision boundary, Sensitive attributes, Unprivileged groups, Classification},
            pdfproducer={LaTeX},
            pdfcreator={pdflatex}]{hyperref}
\usepackage{xcolor}
\usepackage{graphicx}
\usepackage{subcaption}
\usepackage{amsmath}
\usepackage{amssymb}
\usepackage{cite}
\usepackage{booktabs}
\usepackage{tabularx}
\usepackage{verbatim}


\newglossaryentry{Burden}{
    name=Burden,
    first={\emph{Burden}},
    description={A term we use for burden}
}
\newglossaryentry{statpar}{
    name=SP,
    first={\emph{statistical parity} (SP)},
    plural=statistical parities,
    description={}
}
\newglossaryentry{dempar}{
    name=demographic parity,
    first={\emph{demographic parity}},
    plural=demographic parities,
    description={}
}
\newglossaryentry{recourse}{
    name=recourse,
    first={\emph{recourse}},
    description={}
}
\newglossaryentry{sharma}{
    name=Sharma et al.~\cite{certifai},
    first={Sharma, Henderson, and Ghosh~\cite{certifai}},
    description={}
}
\newglossaryentry{legitimate}{
    name=legitimate,
    first={legitimate (non-sensitive and non-proxying)},
    description={}
}

\begin{document}

\title{Bursting the Burden Bubble?}
\subtitle{An Assessment of Sharma et al.'s Counterfactual-Based Fairness Metric}

\author{
Yochem van Rosmalen
\and
Florian van der Steen
\and
Sebastiaan Jans
\and
Daan van der Weijden
}

\authorrunning{Y. van Rosmalen et al.}

\institute{
Utrecht University, Heidelberglaan 8, 3584 CS Utrecht, The Netherlands\\
\email{\{y.m.vanrosmalen,f.a.vandersteen,s.j.j.jans,d.j.vanderweijden\}@students.uu.nl}}

\maketitle

\begin{abstract}
Machine learning has seen an increase in negative publicity in recent years, due to biased, unfair, and uninterpretable models. There is a rising interest in making machine learning models more fair for unprivileged communities, such as women or people of color. Metrics are needed to evaluate the fairness of a model. A novel metric for evaluating fairness between groups is Burden, which uses counterfactuals to approximate the average distance of negatively classified individuals in a group to the decision boundary of the model. The goal of this study is to compare Burden to statistical parity, a well-known fairness metric, and discover Burden's advantages and disadvantages. We do this by calculating the Burden and statistical parity of a sensitive attribute in three datasets: two synthetic datasets are created to display differences between the two metrics, and one real-world dataset is used. We show that Burden can show unfairness where statistical parity can not, and that the two metrics can even disagree on which group is treated unfairly. We conclude that Burden is a valuable metric, but does not replace statistical parity: it rather is valuable to use both.
\end{abstract}

\section{Introduction}

Automated decision making has been used in many real-world applications, e.g. loan applications and predicting recidivism of criminals \cite{loan,skeem}. However, many of these algorithms are black boxes, and their decision processes are not transparent to humans. This is undesirable since it could lead to the unfair treatment of certain groups, without being able to provide an explanation \cite{angwin2016machine}. This has led to an increased demand of fair and explainable models, with many new frameworks for providing explanations being proposed \cite{lundberg2017shap,ribeiro2016lime,ribeiro2018anchors}, as well as new metrics to measure the fairness of a model \cite{kamiran2009demographicparity,hardt2016equalisedodds,woodworth2017learning}.

Metrics for fair machine learning measure how well a particular model is towards different groups within a dataset. Although the definition and practical implementation of fairness varies between different metrics, their overarching goal is to provide insight into the level of fairness between different groups regarding sensitive attributes (e.g. age, gender, socioeconomic status).

One of the new frameworks is CERTIFAI \cite{certifai}, a framework that tests the robustness of a model, as well as providing explanations and a metric to measure fairness. This framework is implemented commercially by the company CognitiveScale, and used by many organizations in different domains.

It does so by generating counterfactuals\footnote{The term \textit{counterfactual} in this context does not refer to the  type of counterfactual discussed in literature on causality.} for each datapoint in the dataset. This counterfactual is a synthetic datapoint, generated to have the other possible outcome, while being as close as possible to the original datapoint\cite{wachter2017counterfactual}. The counterfactuals provide insight into what features should change to have the model classify the datapoint differently. Not only does this provide an explanation for why a certain classification was made, this also allows us to measure the -- possibly unfair -- difference in treatment for certain groups (e.g. male and female). By calculating the average distance for a group between original datapoints in the negative outcome class (e.g. loan application denied) and their generated counterfactuals (e.g. loan application approved), the \gls{Burden} of a group can be calculated. These \gls{Burden} scores can be compared to see which groups have a higher \gls{Burden} and thus have more difficulty converting from the negative to the positive predicted outcome class. In this way we do not only calculate \textit{if} a group is being disadvantaged, but also \textit{how much} a group is disadvantaged. This can give more detailed insight into the fairness of the model.

\gls{sharma} claim that ``Burden can be considered to be a nuanced version of other fairness measures (such as demographic parity)'' \cite[p.~170]{certifai}.
It is calculated by measuring the ratio of the probability of receiving a positive outcome from a model between groups (See Sec. \ref{sec:sp}). However, this claim of nuance is not validated in their study. In this study, the claim is tested, by comparing the \gls{Burden} metric to \gls{statpar}. We focus on \gls{statpar} because it, like \gls{Burden}, does not take the actual ground truth target value into account but rather the model's prediction. It therefore makes sense to compare the two. 

In this study, we investigate situations where both metrics give different results to see if \gls{Burden} can provide more nuance and if it is a good fairness metric in practice. This is tested on two synthetic datasets with hypothetical data and a real-world loan application dataset \cite{dataset}. All three datasets have a \emph{binary} outcome class: a favorable outcome and an unfavorable outcome. This means that the models used are also binary classification models.

\section{Related Work}\label{sec:relatedwork}

In this section, the fairness metrics \gls{statpar} (Sec.~\ref{sec:sp}) and \gls{Burden} (Sec.~\ref{sec:burden}) are explained more in-depth to get a better theoretical understanding of how and why they work as fairness metrics.

\subsection{Statistical Parity}\label{sec:sp}
There are many metrics to measure how fair a model is, and there is no agreement on a best method, or even on the definition of fairness itself \cite{chouldechova2018frontiers}. However, one of the most common and easy to implement fairness metrics is that of \gls{statpar}, or \gls{dempar} \cite{kamiran2009demographicparity}.
In order to calculate \gls{statpar}, we have to calculate the acceptance rate (AR) for a specific group of a feature $S=s$. For example, if $S$ is a binary value the groups could be 0 and 1, which can be seen in Eq.~\ref{eq:ar}.
\begin{equation}\label{eq:ar}
 AR_{S=s} = P(\hat{Y}=1|S=s)
\end{equation}
This means that the acceptance rate of the group where $S=s$ is defined as the probability ($P$) of the model predicting a positive outcome ($\hat{Y}=1$), given that $S=s$.\footnote{The actual ground truth outcome, or target value, of a supervised dataset is denoted as $Y$, while the model's predicted outcome is denoted as $\hat{Y}$.} To calculate the \gls{statpar} between two groups of a binary feature $S$, we look at the ratio of the acceptance rate of both groups, as seen in Eq.~\ref{eq:sp}.
\begin{equation}\label{eq:sp}
    SP_{S} = \frac{AR_{S=0}}{AR_{S=1}} = \frac{P(\hat{Y}=1|S=0)}{P(\hat{Y}=1|S=1)} 
\end{equation}
If the same percentage of individuals receives a positive score for each group, and thus the outcome of the ratio is 1, the two groups both have the same probability of receiving a positive outcome prediction from the model. This is seen as fair: if $S$ is a sensitive attribute, e.g. age, there should be no difference in receiving a positive prediction. Perfect SP is almost never possible in practice, so often the 80\% rule for disparate impact \cite{feldman2015certifying} is used: there is disparate impact if $SP \leq 0.8$.

\subsection{Burden}\label{sec:burden}

The CERTIFAI framework \cite{certifai} uses counterfactuals to measure different treatment of groups. A counterfactual, in this context, is a datapoint calculated to be as similar to an original datapoint as possible, while receiving a different classification. A counterfactual datapoint can provide an individual with \gls{recourse}: 
the counterfactual datapoint can show the individual which changes to the input features are can be made to change the classification to the desired output class. To this end, counterfactuals are also constrained to be realistically achievable. It is, for example, not useful to find a counterfactual with different gender, since this is not something one will realistically change to achieve a different classification.

\begin{figure}
    \centering
    \includegraphics[width=\textwidth]{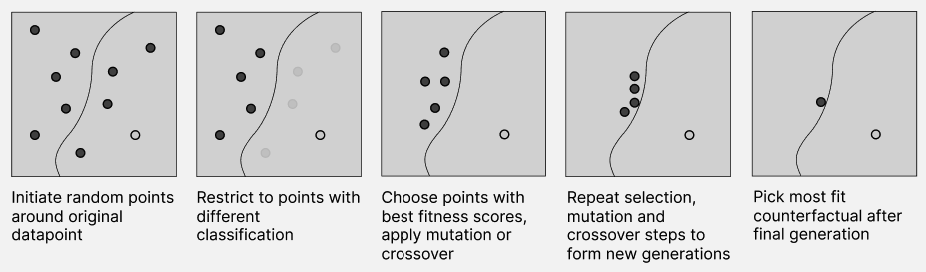}
    \caption{Visual representation of the counterfactual generation process of CERTIFAI. Adopted from \cite{certifai}.}
    \label{fig:cf-generation}
\end{figure}

\Gls{sharma} propose a genetic heuristic search for generating a counterfactual $c$ for a datapoint $x$. The process is shown visually in Fig.~\ref{fig:cf-generation}. Starting from a randomly initialised population of size $N$, the counterfactuals $\mathbf{c}$ that are classified in the opposing class are selected. These are then mutated with probability $P_m$, which involves arbitrarily changing some feature values. Subsequently, crossover is applied with probability $P_c$, which involves randomly interchanging some feature values between individuals. Then, a top-$k$ selection procedure is applied where only the most fit counterfactuals are selected. The fitness function $\frac{1}{d(x,c)}$ is the inverse of a distance function calculated over a datapoint and its counterfactual. The population is then filled back up to $N$ by randomly generating new counterfactual points. This process is repeated for a predetermined maximum of generations. Finally, the fittest counterfactual $\mathbf{c^*}$ is selected for each datapoint.

Using $\mathbf{c^*}$, the \gls{Burden} of a group can be calculated. The Burden of a group with value $s$ for feature $S$ is calculated over the instances with value $s=S$ that are classified in the unfavorable class. Burden is then defined as the mean of the distances between these datapoints and their counterfactuals,
\begin{equation}\label{eq:burden}
    \text{Burden}_{S=s} = \mathbb{E}_{S=s}[d(\mathbf{x}, \mathbf{c^*})],
\end{equation}
where the distance function can be chosen and corresponds to the distance function in the fitness calculation. Equation \ref{eq:burden} corresponds to equation 11 in \cite{certifai}. Similarly to \gls{statpar}, we can use the ratio of two \gls{Burden} scores for a binary sensitive feature. 

\section{Methods}\label{sec:methods}
The methodology is broken down in four parts: the creation of the two synthetic datasets, the description of the `Default of Credit Card Clients' dataset, the classifier models and lastly CERTIFAI's counterfactuals and Burden. The Python code (using Jupyter Notebook), saved models, and generated data is available on GitHub\footnote{\url{https://github.com/yochem/bursting-the-burden-bubble}}.

\subsection{Synthetic Datasets}\label{sec:syndata}

We created two datasets to demonstrate two types of disagreements between \gls{Burden} and \gls{statpar} that are theoretically possible. One synthetic dataset, $D_A$, shows that \gls{Burden} disagrees with \gls{statpar} about \textit{whether there is} unfairness, and the other dataset, $D_B$, shows that \gls{Burden} disagrees with \gls{statpar} about \emph{which group} is treated unfairly. Both synthetic datasets consist of 80 datapoints.

Each datapoint in the two datasets consists of three features and a label. The \gls{legitimate}\footnote{A proxying feature can reveal sensitive information. E.g. someone's address, although not a sensitive feature, can reveal someone's socioeconomic status because of their neighborhood.} features $X_1$ and 
$X_2$, the sensitive attribute $S$ (0 is unprivileged, 1 is privileged), and the target label $Y$ (0 is unfavorable, 1 is favorable). This means that datapoint $i$ is given by 
$ D^{(i)} = (x_1, x_2, s, y)$. The \gls{legitimate} features $X_1$ and $X_2$ are mixtures of Gaussians around multiple means $\mu$. For example, $X_1^{(i)} \sim \mathcal{N}(\mu_1^{(i)}, \sigma)$ means that the feature $X_1$ from point $i$ is sampled from a normal distribution with mean $\mu_1^{(i)}$. The different $\mu$-values, along with the values of the other features, are listed in Table~\ref{tab:syndata}. All samples have a standard deviation $\sigma$ of 1. The sensitive attribute $S$ is selected (not sampled from a random distribution), as is the target label $Y$. The true underlying function between the legitimate features and the outcome can be derived from Table~\ref{tab:syndata}. The datapoints of the two datasets are plotted in Fig.~\ref{fig:syndataplot}.

\subsubsection{Dataset on Presence of Unfairness $D_A$}
\gls{Burden} takes the average distance of a group to their counterfactuals into account, while \gls{statpar} does not. 
Therefore, the synthetic data needs to satisfy two properties: Firstly, it needs to satisfy \gls{statpar}, so for each group, the same number of datapoints needs to be predicted positive.
Secondly, the average distance of the negatively predicted datapoints to their counterfactuals needs to differ between the two groups to show how \gls{Burden} can find this unfairness.

\subsubsection{Dataset on Direction of Unfairness $D_B$}
This dataset should let \gls{Burden} and \gls{statpar} disagree on which group is treated unfairly. This means that \gls{statpar} has to label one group of the sensitive attribute as unprivileged, and \gls{Burden} has to label the other group as unprivileged. \Gls{statpar} labels a group as unprivileged if the group has fewer  positively predicted outcomes than the other group ($P(\hat{Y}=1|S=0) \neq P(\hat{Y}=1|S=1)$). With a perfect classifier (accuracy of 1), we have $\hat{Y} = Y$.
Using the definition of \gls{statpar}, an unprivileged group can be formed by having relatively fewer datapoints where $\hat{Y}=1$. For \gls{Burden} to disagree with \gls{statpar}, the other $S$-group (i.e. the group that \gls{statpar} sees as privileged) needs to have a greater distance to their counterfactuals at the decision boundary, as illustrated in Fig. \ref{fig:cf-generation}.

\begin{table}
    \centering
    \caption{The distribution of values per feature for both datasets $D_A$ and $D_B$. The count is the number of datapoints sampled from the normal distributions for $X_1$, and $X_2$, with shown mean $\mu$ for their normal distribution $\mathcal{N}(\mu, 1)$.}
    \label{tab:syndata}
    \begin{tabularx}{.7\textwidth}{@{\extracolsep{\fill}}cccccccccc}
        \toprule
        \multicolumn{5}{c}{Dataset $D_A$} & \multicolumn{5}{c}{Dataset $D_B$}\\
        \cmidrule(lr){1-5}\cmidrule(lr){6-10}
        $\mu_{X_1}$ & $\mu_{X_2}$ & $S$ & $Y$ & count & $\mu_{X_1}$ & $\mu_{X_2}$ & $S$ & $Y$ & count\\
        \midrule
        1 & 9 & 0 & 0 & 20 & \hspace{0.2cm} 1 & 9 & 1 & 0 & 15\\
        3.5 & 5 & 1 & 0 & 20 &\hspace{0.2cm} 3.5 & 5 & 0 & 0 & 15\\
        9 & 1 & 0 & 1 & 20 &\hspace{0.2cm}  9 & 1 & 1 & 1 & 30\\
        9 & 1 & 1 & 1 & 20 & \hspace{0.2cm} 9 & 1 & 0 & 1 & 20\\
        \bottomrule
    \end{tabularx}
\end{table}

\subsection{Default of Credit Card Clients Dataset}\label{sec:taiwan}

To explore the claim of \gls{Burden} being more nuanced, the metric is also compared to \gls{statpar} on real-world data. This is done on a subset of the Default of Credit Card Clients dataset, also known as the Taiwan loan dataset, from \cite{dataset} (from now on called Taiwan dataset). This is a dataset of credit card users, which records whether the individual defaults on a loan. Since defaulting on a loan is a negative outcome, the favorable label in this dataset is 0: `did not default'. The unfavorable label is 1: `default'. The sensitive attributes are gender, education, marriage, and age \cite{eusex}. These are not used as training input data. After training, we use the sensitive attribute gender for testing the fairness of the model. Monthly payments were tracked for the other features, such as history of past payment, amount of bill statement, amount of previous payment, and amount of given credit. Datapoints with values not following the specification\footnote{An example: gender is encoded as either 1 or 2; some datapoints had a value of 3 in the column of gender. Since this value is not described in the dataset specification, the datapoints where gender does not equal 1 or 2 were removed.} were removed. The dataset contains 30,000 instances. The computational cost for generating counterfactuals is large because the genetic algorithm iteratively goes over large population sizes for many generations. Therefore, this study is limited to a random sample of 1000 instances from the Taiwan dataset.

\subsection{Classifier}\label{sec:model}
A binary classification model (classifier) is needed for calculating \gls{Burden}, and \gls{statpar}. On all three datasets, a Logistic Regression model \cite{cox1958regression} was trained. The Logistic Regression model was chosen because of its simplicity and the interpretability of its linear decision boundary.

The classifiers were trained on the datasets without their sensitive features. No hyper-parameter optimization was performed, and the datasets were not partitioned into separate tests for training, and evaluation. This was done to remove unnecessary complexity: our interest lies in evaluating fairness, not model performance. The Logistic Regression model was implemented in PyTorch for reasons concerning compatibility with the CERTIFAI framework. It used the binary cross-entropy loss function \cite{cox1958regression} and the stochastic gradient descent optimizer \cite{robbins1951stochastic}. The learning rate was 0.001 and the number of iterations was 2000. The number of input dimensions for each classifier was the number of \gls{legitimate} features, i.e. $X_1$ and $X_2$ for the synthetic datasets, and 19 features for the Taiwan dataset relating to past payments, bill statements, and credit features.

\subsection{CERTIFAI's Burden}\label{sec:certifai}
Using the \texttt{CERTIFAI.fit()} method, counterfactuals were generated given the model. The hyperparameters were the following: 10 generations of populations with size 60,000, of which at most 10,000 are retained after selection, of which at most 5,000 are retained for the next generation, and unconstrained generating of counterfactual features. The probabilities for crossover and mutation were adopted from \cite{certifai}. For calculating the Burden, CERTIFAI's \texttt{check\_fairness} method was used with as argument a mapping containing 1) the 
sensitive attribute and its value (e.g. \texttt{s: 0}), and 2) that it should be calculated over the unfavorable class (i.e. \texttt{favorable: 0}).

\section{Results}\label{sec:results}

\begin{figure}
    \centering
    \begin{subfigure}{0.49\textwidth}
        \includegraphics[width=\textwidth]{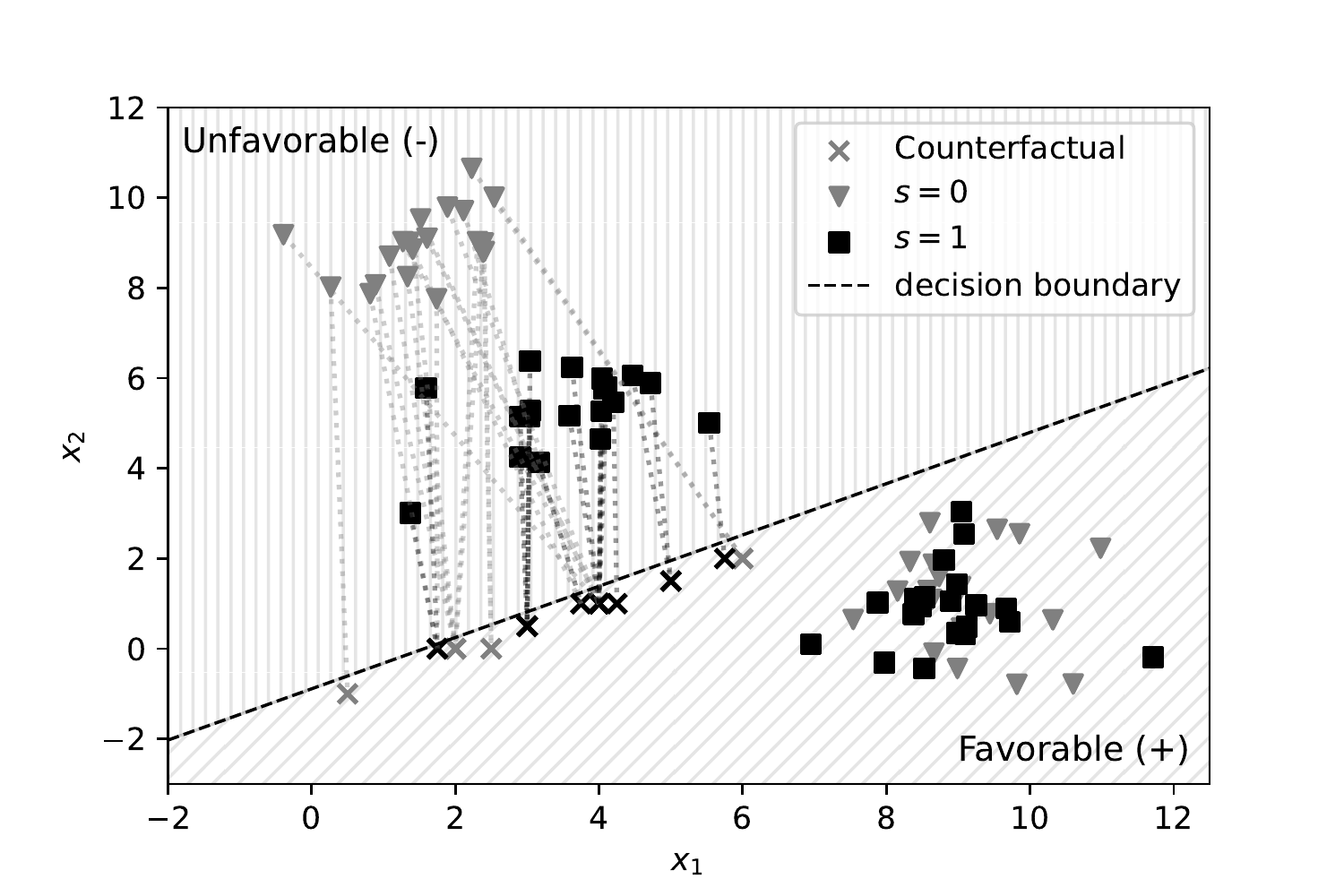}
        \caption{$D_A$, where \gls{Burden} and \gls{statpar} disagree on the presence of unfairness.}
        \label{fig:syndatafavor}
    \end{subfigure}
    \begin{subfigure}{0.49\textwidth}
        \includegraphics[width=\textwidth]{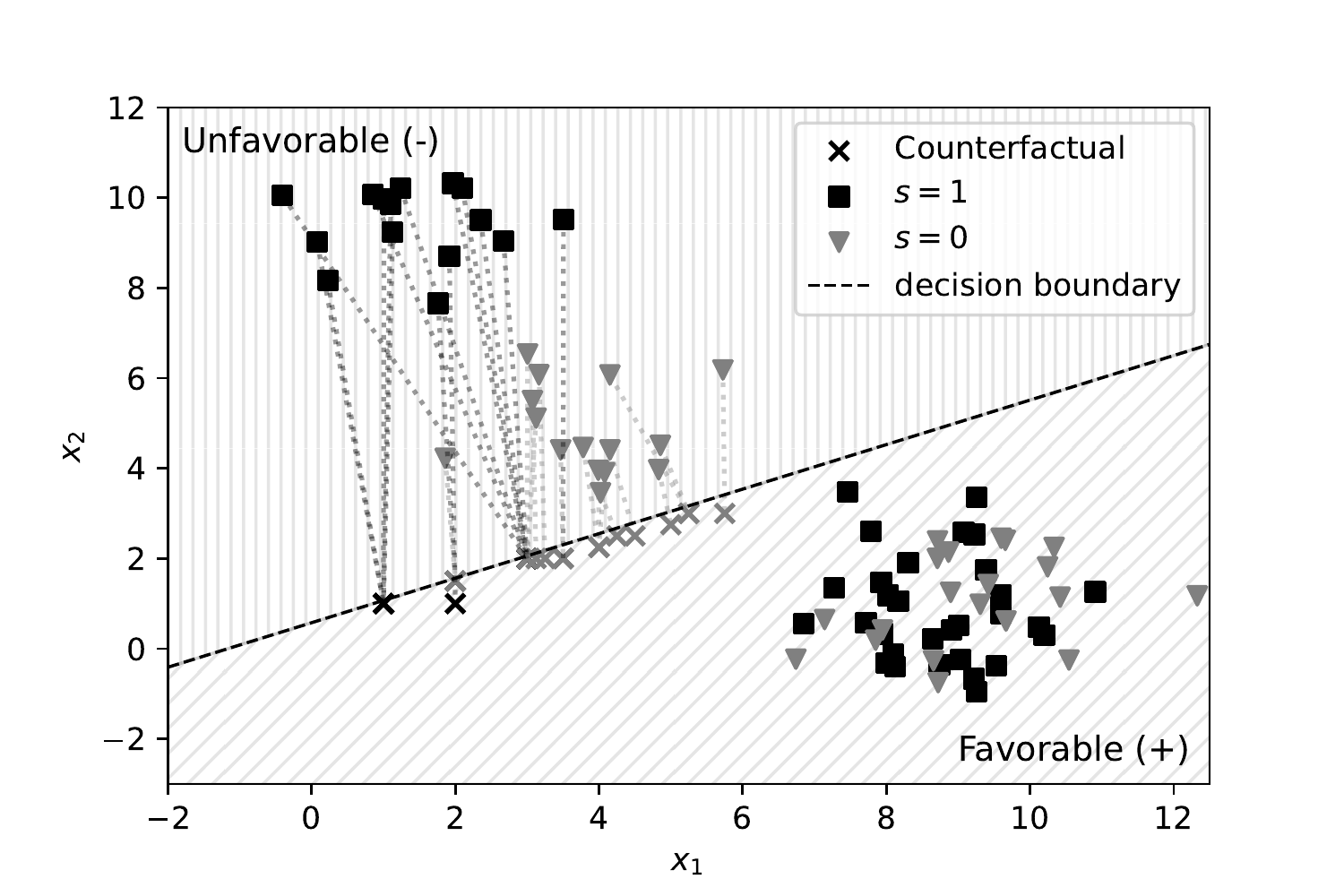}
        \caption{$D_B$, where \gls{Burden} and \gls{statpar} disagree on the direction of unfairness.}
        \label{fig:syndataunfavor}
    \end{subfigure}
    \caption{The synthetic datapoints ($\blacktriangledown,\blacksquare$) for datasets $D_A$ and $D_B$. The counterfactuals (\texttimes) for datapoints from the unfavorable outcome class are also included and are connected using a dotted line. The decision boundary (\texttt{---}) of the classifier is also shown. Note that the distribution of the positive class does not influence \gls{Burden}.}
    \label{fig:syndataplot}
\end{figure}

In the first two experiments, logistic regression models were trained on $D_A$ and $D_B$ respectively and both got an accuracy of 1.00. In Fig. \ref{fig:syndataplot}, this is shown by the decision boundaries laying perfectly between both groups. After the models were trained, the counterfactuals for the unfavorable class were generated by CERTIFAI, also shown in Fig. \ref{fig:syndataplot}. After this, the \gls{statpar} and \gls{Burden} were calculated. The results of the metrics on both experiments are listed in Table \ref{table:results}.

For the experiment on $D_A$ we see that \gls{statpar} is met: the ratio of the acceptance rates of the groups is 1. The Burden of group $S=0$ is higher (Burden of 11.6) than of group $S=1$ (Burden of 4.65). This difference in Burden can also be eyeballed using Fig.~\ref{fig:syndatafavor}, where the $S=0$ group is further away from their counterfactuals than the $S=1$ group.

For the experiment on $D_B$ we see that \gls{statpar} is 0.857. The Burden of the two groups are 3.31 and 11.0 for $S=0$ and $S=1$, respectively. The datapoints are plotted in Fig.~\ref{fig:syndataunfavor}.

The results of the last experiment, on the Taiwan dataset, are also listed in Table \ref{table:results}. The model trained on this dataset achieved an accuracy of 0.78. The results are the following: \Gls{statpar} is nearly met, with a value of 1.02 (0.967 over 0.948). \Gls{Burden} however shows that females have almost 1.5 times higher Burden than males, respectively 1.38 and 0.940.

\begin{table}
    \centering
    \caption{\Gls{statpar} and \gls{Burden} for the three datasets. The acceptance rate and \gls{Burden} are given per group ($S=0$ and $S=1$ for the synthetic datasets correspond to gender=female and gender=male respectively for the Taiwan dataset), as well as the \gls{Burden} ratio and statistical parity (SP) between the two groups, in bold.}
    \label{table:results}
    \begin{tabularx}{.8\textwidth}{l@{\extracolsep{\fill}}cccccc}
    \toprule
    & \multicolumn{2}{c}{Acceptance Rate} & SP & \multicolumn{3}{c}{Burden}\\
    \cmidrule(lr){2-3}\cmidrule(lr){4-4}\cmidrule(lr){5-7}
    Dataset & $S=0$ & $S=1$ & 0/1 & $S=0$ & $S=1$ & 0/1\\
    \midrule
    $D_A$  & 0.500 & 0.500 & \textbf{1.00} & 11.6 & 4.65 & \textbf{2.49}\\
    $D_B$  & 0.571 & 0.667 & \textbf{0.857} & 3.31 & 11.0 & \textbf{0.302}\\
    Taiwan & 0.967 & 0.948 & \textbf{1.02} & 1.38 & 0.940 & \textbf{1.47}\\
    \bottomrule
    \end{tabularx}
\end{table}

\section{Discussion}\label{sec:discussion}

In this section, the results are discussed as well as the limitations of this study and directions for future work.

\subsection{Discussion on Experimental Results}
In the first experiment with dataset $D_A$, the results show that even though \gls{statpar} was met (ratio of 1.00), Burden shows that the model is unfair towards group $S=0$, since their Burden is higher. This means that Burden can show unfairness between groups when \gls{statpar} can not. This is a positive result for the claim of \gls{sharma} that \gls{Burden} provides more nuance than \gls{statpar} in this situation. The distance to the counterfactuals near the decision boundary is important here to actually find the unfairness.

In the second experiment on dataset $D_B$, the \gls{statpar} shows unfairness towards group $S=0$ since \gls{statpar} is less than 1. Burden however tells us that group $S=1$ is being treated unfairly, because the Burden of this group (11.0) is higher than the Burden of the other group (3.31). This means that \gls{statpar} and Burden can disagree on which group is treated unfairly.

As a third experiment, on real-world Taiwan data, the results show the same effect as the first experiment. Although the difference is smaller than with $D_A$, the results show that with \gls{statpar} the model is more fair than with Burden. Burden can thus add nuance compared to \gls{statpar} in this situation.

\subsection{Limitations and Future Work}
Future work could look for real-world examples of the synthetic dataset $D_B$, and if the disagreement between \gls{Burden} and \gls{statpar} found in our synthetic experiment occurs in other situations.

Furthermore, it is important to note that the computational complexity of the \gls{Burden} metric is extremely high in comparison to a metric like \gls{statpar}. While \gls{statpar} is a simple calculation of a ratio between two percentages, the calculation of a single counterfactual for the \gls{Burden} metric can take minutes. For large datasets it might thus be necessary to compute this metric for a representative sample of the dataset.

A limitation of the experiment performed on the Taiwan dataset is that it is hard to verify the quality of the counterfactuals. The complexity of the decision boundary and the dimensionality of the data hinders the visualization of the counterfactuals. Any found difference in \gls{Burden} on this dataset might thus be a measurement error. Another limitation is that the experiment was performed on a small part of the data, as the calculation of 30000 counterfactuals takes a very large amount of resources.

\section{Conclusion}\label{sec:conclusion}
In this study we assessed the fairness metric introduced in \gls{sharma}, using three experiments.
The first experiment, using synthetic dataset $D_A$, shows that \gls{Burden} can pick up unfairness when \gls{statpar} can not. The second experiment, using synthetic dataset $D_B$, shows that \gls{Burden} and \gls{statpar} can even disagree on which group is treated unfairly. The last experiment, using the Taiwan dataset, shows that \gls{Burden} is more nuanced than \gls{statpar} on a real-world dataset. The three experiments show that \gls{Burden} has the ability to provide more information than \gls{statpar}, but this information may not be in line with \gls{statpar}.

We conclude that \gls{Burden} and \gls{statpar} can be complementary, as both metrics measure different, important aspects of model fairness. However, due to the computational complexity of the evolutionary algorithm, \gls{Burden}'s use of recourse might not outweigh the speed of \gls{statpar} for all models. For simple models, where the decision boundary can be easily computed, a genetic algorithm might be unnecessarily complicated. We agree with \gls{sharma} that \gls{Burden} can provide more nuance than \gls{statpar}, but the increase in nuance might not always be worth the computational cost.

\subsubsection*{Acknowledgments.} We thank dr. Dong Nguyen, Yupei Du, and dr. Heysem Kaya for their great help and this opportunity.

\bibliographystyle{splncs04}
\bibliography{references}

\end{document}